\documentclass[journal]{IEEEtran}
\usepackage{amsmath,amsfonts}
\usepackage{algorithmic}
\usepackage{algorithm}
\usepackage{array}
\usepackage[caption=false,font=normalsize,labelfont=sf,textfont=sf]{subfig}
\usepackage{textcomp}
\usepackage{stfloats}
\usepackage{url}
\usepackage{verbatim}
\usepackage{graphicx}
\usepackage{cite}
\usepackage{booktabs}
\begin{document}

\title{MLLM-Guided Semantic Correction for Text-to-Video Generation}

\author{Junhao Chen, Zheqi Lv, Keting Yin, Shengyu Zhang, \\Zhou Zhao, Feiyang Chen, Xinyu Duan, Baoxing Huai, and Fei Wu
\thanks{Junhao Chen and Keting Yin are with the School of Software Technology, Zhejiang University, Hangzhou 310027, China (e-mail: chenjunhao100@zju.edu.cn; yinkt@zju.edu.cn).}
\thanks{Zheqi Lv, Shengyu Zhang, Zhou Zhao, and Fei Wu are with the College of Computer Science and Technology, Zhejiang University, Hangzhou 310027, China(e-mail: zheqilv@zju.edu.cn; sy\_zhang@zju.edu.cn; zhaozhou@zju.edu.cn; wufei@zju.edu.cn).}
\thanks{Feiyang Chen, Xinyu Duan, and Baoxing Huai are with the AI System Innovation Lab, Huawei Cloud Computing Technology Co., Ltd., Hangzhou 310005, China(e-mail: chenfeiyang2@huawei.com; duanxinyu@huawei.com; huaibaoxing@huawei.com).}

}



\maketitle

\begin{abstract}

Recent advances in diffusion models and Transformer architectures have led to significant progress in text-to-video generation. However, these models often suffer from semantic errors such as missing objects, incorrect attributes, or mismatched actions. Although some semantic correction methods perform optimization before sampling or refinement after sampling, how to detect and correct semantic deviations during the video generation process remains underexplored.
In this paper, we introduce a training-free, interpretable mid-generation correction framework that integrates multimodal large language model (MLLM) feedback directly into the diffusion sampling loop. Our framework achieves diffusion trajectory correction by injecting semantic evaluation signals during video synthesis, enabling the model to optimize the generated content through continuous self-reflection. We propose two key modules: a Semantic Assessment Supervisor that generates intermediate preview frames for semantic evaluations and deviation diagnostics, and a Semantic Modification Assistant that corrects semantic drift during inference via a controllable latent trajectory intervention. Our method improves semantic alignment, visual fidelity, and temporal consistency without modifying model parameters. We validate the effectiveness of our approach through extensive experiments across multiple benchmarks.

\end{abstract}

\section{Introduction}
\label{sec:intro}

\IEEEPARstart{R}ecently, rapid advances in diffusion models~\cite{ho2020denoising, song2021scorebased} and transformer architectures~\cite{peebles2023scalable} have resulted in substantial progress in the field of text-to-video generation.
Both proprietary business models (e.g., Sora~\cite{openai2024Sora}) and open-source frameworks (e.g., HunyuanVideo~\cite{kong2024hunyuanvideo}, CogVideoX~\cite{hong2022cogvideo, yang2024cogvideox}) exhibit a strong ability to adhere to textual prompts throughout the generation process, substantially improving overall video quality.
However, conventional text-to-video diffusion models typically lack mechanisms for understanding the semantics encoded in the intermediate latent representations. During the generation process, the lack of semantic awareness prevents the model from detecting and correcting semantic deviations, such as missing objects, incorrect attributes, or mismatched actions (see Fig.~\ref{fig:case}).

\begin{figure}[t]
  \centering
    \includegraphics[width=1\linewidth]{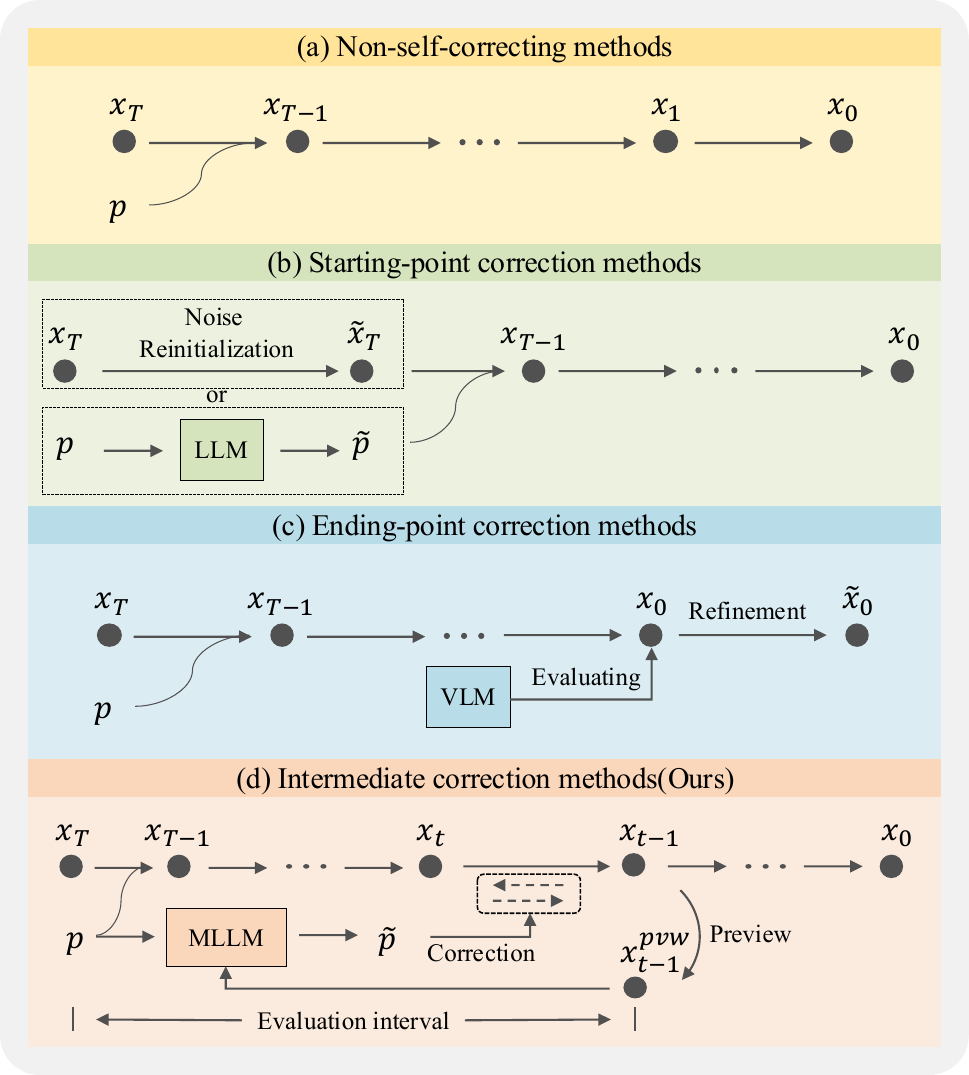}
    \caption{
        \textbf{Overview of semantic correction strategies for text-to-video generation.}
        (a) \textit{Non–self-correcting}: fixed-scale guidance tends to amplify artifacts and motion errors.
        (b) \textit{Starting-point correction}: prompt/noise optimization before sampling cannot prevent drift during generation.
        (c) \textit{End-point correction}: post-generation refinement reacts to errors but cannot prevent them.
        (d) \textit{Intermediate correction(Ours)}: a training-free, MLLM-guided mid-generation correction that injects semantic feedback into the diffusion loop to maintain coherence during synthesis.
    }
    \label{fig:existing_solutions}
\end{figure}

To address these challenges, existing approaches can be broadly categorized into three groups (see Fig.~\ref{fig:existing_solutions}).
(1) \textit{Non–self-correcting methods}: Classical techniques such as Classifier-Free Guidance (CFG)~\cite{ho2022classifierfreediffusionguidance} enhance prompt adherence by scaling the guidance strength. However, excessively high and static guidance scales often compromise visual fidelity and motion diversity while amplifying accumulated errors along the diffusion trajectory.
(2) \textit{Starting-point correction methods}: Methods (e.g., Free-Bloom~\cite{huang2023free}, GPT4Motion~\cite{lv2024gpt4motion}, FreeInit~\cite{wu2024freeinit}) improve alignment by optimizing prompts or the initial noise distribution before sampling. These techniques mitigate early-stage drift but are inherently unable to react to semantic deviations that occur later in the trajectory.
(3) \textit{End-point correction methods}: Approaches (e.g., VideoRepair~\cite{lee2025videorepairimprovingtexttovideogeneration}, NeuS-E~\cite{choi2025wellfixpostimproving}) perform refinement after the generation is completed or during the late inference stage by employing external scorers or feedback evaluators to assess and correct semantic inconsistencies. Although these refinement strategies improve semantic alignment, they are inherently reactive, correcting errors post-generation instead of preventing them during sampling. 
In summary, existing methods lack the capacity to adaptively correct dynamic semantic deviations within the diffusion sampling process.

In this paper, we introduce a \textit{training-free, interpretable, midgeneration correction} framework that integrates multimodal large language model (MLLM) feedback directly into the diffusion sampling loop. The framework injects semantic evaluation signals during video synthesis, enabling online trajectory correction. Our key insight is to enable the diffusion model to operate analogously to a painter who inspects and adjusts the evolving canvas, thereby functioning as an active semantic reasoner that performs continuous self-reflection and refinement during generation. Specifically, an MLLM is employed to analyze intermediate generations and guide the diffusion process to correct semantic drift, introducing an external supervisory signal that prevents error accumulation during sampling. 

Our insights guide our progressive pipeline, which consists of two modules: \textit{Semantic Assessment Supervisor} and \textit{Semantic Modification Assistant}. (i) \textit{Semantic Assessment Supervisor}. A naive strategy of providing raw, noisy latents to an MLLM for evaluation proves ineffective, as intermediate representations lack coherent semantics. Therefore, at selected sampling steps, the Semantic Assessment Supervisor makes the model generate intermediate preview frames that are processed by the MLLM to produce semantic evaluations and bias diagnostics. (ii) \textit{Semantic Modification Assistant}. Once these assessments are available, a central technical question remains: How to incorporate feedback into the diffusion trajectory in a controllable and temporally consistent manner without violating model priors? To solve this, we introduce a Semantic Modification Assistant. First, we perform an unconditional reverse diffusion to roll the current latent back to an unconditional anchor state, thereby removing accumulated conditional errors. Then, using MLLM-derived enhancement and suppression cues, a conditional forward diffusion step reintroduces corrected semantic constraints into the generation process. Together, the \textit{Semantic Assessment Supervisor} converts the diffusion model from a passive generator into an active self-monitor, while the \textit{Semantic Modification Assistant} enables dynamic, inference-time semantic correction without modifying generator parameters, which we term understanding-aware generation.

The contributions of this paper can be summarized as:
\begin{itemize}
    \item We propose a \textit{training-free,  interpretable, mid-generation correction} framework that embeds MLLM feedback directly into the text-to-video diffusion sampling loop, improving semantic alignment and visual quality without modifying model parameters.
    \item We design two key modules, \textit{Semantic Assessment Supervisor} and \textit{Semantic Modification Assistant}, that respectively enable reliable interpretation of intermediate states and controllable corrective interventions within the ongoing sampling process, achieving inference-time semantic correction.
    \item We demonstrate through extensive experiments that our method significantly enhances semantic consistency, visual fidelity, and temporal coherence across multiple text-to-video benchmarks, validating the efficacy of MLLM-guided semantic feedback during diffusion sampling.
\end{itemize}

\section{Related Work}
\label{sec:related}

\subsection{Text-to-Video Generation}
Text-to-video (T2V) generation aims at synthesizing coherent videos conditioned on textual descriptions. Recent advances in diffusion models have established two main research paradigms: training-based~\cite{an2023latentshiftlatentdiffusiontemporal,blattmann2023align,esser2023structure,ge2023preserve,ho2022imagen,ho2022video,singer2022make,wang2023modelscope,yin2023nuwa,zhang2025show,zhou2022magicvideo, chen2020scripted, gao2025ca2, zhao2024magdiff, Li_2025_ICCV} and training-free approaches~\cite{huang2023free,wu2024freeinit,khachatryan2023text2video, hyun2023frequency}. 
Training-based methods typically train large-scale video diffusion models directly on paired video–text datasets to capture spatial and temporal correlations. For example, Video Diffusion Models (VDM)~\cite{ho2022video} extend image diffusion UNets into 3D UNets for joint spatio-temporal learning, while Make-A-Video~\cite{singer2022make} learns visual–textual alignment from image–text pairs and acquires motion understanding from unlabeled video data. These methods achieve high visual fidelity but require substantial computational resources and large video–text datasets for training.
In contrast, training-free methods leverage pretrained text-to-image diffusion models to synthesize videos without additional training. Approaches such as Text2Video-Zero~\cite{khachatryan2023text2video}, FreeBloom~\cite{huang2023free}, and FreeInit~\cite{wu2024freeinit} adapt image diffusion backbones by introducing temporal attention, cross-frame consistency modules, or motion priors to produce temporally coherent frames. Despite their efficiency, these models often suffer from semantic deviations, generating scenes or actions that drift from the input text due to the lack of explicit semantic reasoning.
To address this issue, we propose a training-free MLLM-guided inference framework that integrates textual and visual understanding into the diffusion process.

\subsection{LLM-assisted Video Generation}
Recent advances in Large Multimodal Models (LMMs)~\cite{achiam2023gpt,anil2023palm,workshop2022bloom} have significantly influenced video generation. Several works employ Large Language Models (LLMs) as high-level planners that interpret textual prompts into structured scene descriptions or multiscene scripts before generation~\cite{lin2023videodirectorgpt,lian2023llm,hong2023direct2v,huang2023free,lv2024gpt4motion,long2024videodrafter,yuan2024mora,lee2025videorepairimprovingtexttovideogeneration,choi2025wellfixpostimproving}. 
For example, LVD~\cite{lian2023llm} transforms text into detailed scene layouts to guide diffusion-based video synthesis, while VideoDrafter~\cite{long2024videodrafter} employs LLMs to produce multiscene scripts and maintain entity consistency via reference images. 
Similarly, Mora~\cite{yuan2024mora} introduces an LLM-driven multi-agent framework for generalist video generation and editing.
While these methods apply LLM reasoning as static guidance before or after diffusion, our mid-generation MLLM mechanism provides dynamic semantic feedback, which relates to recent efforts that leverage semantic representations for perceptual evaluation and content understanding~\cite{chen2024topiq, li2025misc, zhang2024task}.

\section{Method}
\label{sec:method}

\begin{figure*}
    \centering
    \includegraphics[width=\linewidth]{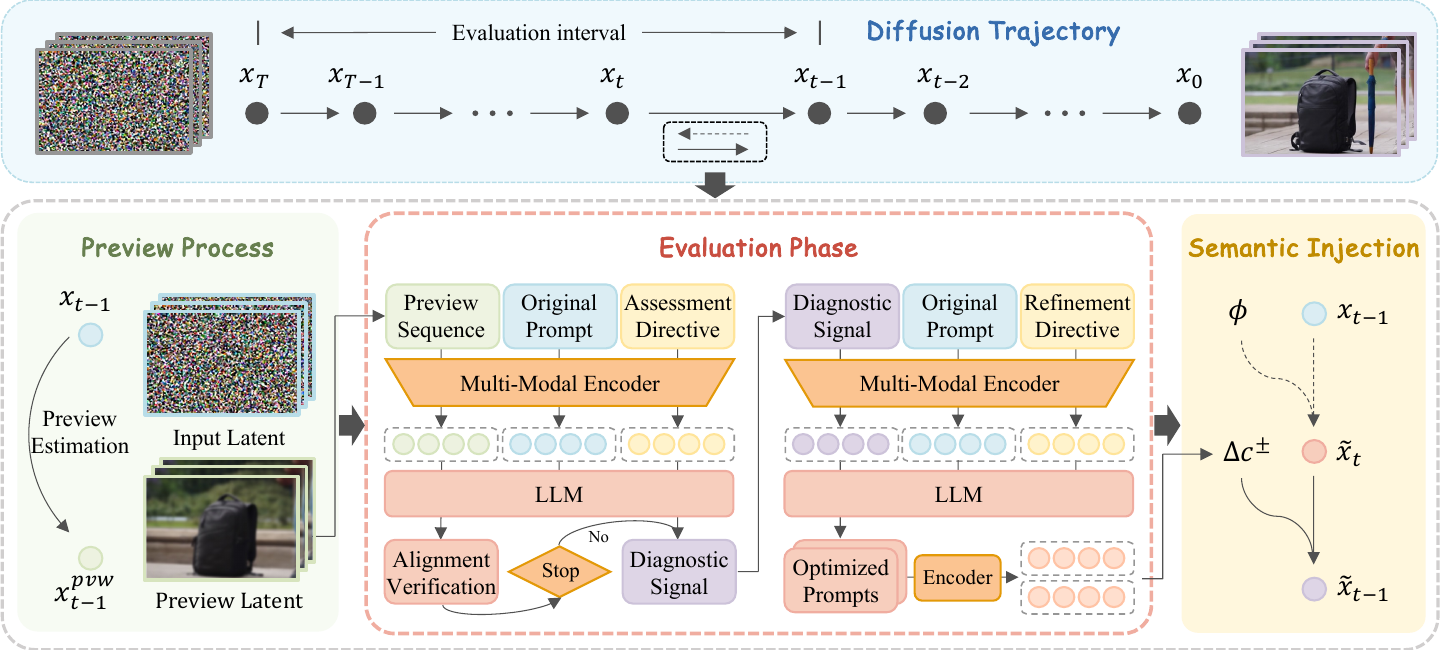}
    \vspace{-0.6cm}
    \caption{
    \textbf{Overview of the proposed mid-generation semantic correction framework (SASMA).}
    The framework, termed SASMA (Semantic Assessment Supervisor and Modification Assistant),  integrates MLLM feedback into the diffusion sampling loop for dynamic semantic correction during text-to-video generation. 
    During sampling, intermediate latents are periodically decoded into preview frames for semantic evaluation (left and middle). 
    The MLLM provides diagnostic feedback and corrective prompts, which are encoded into semantic deltas $\Delta c_t^{\pm}$ and injected back into the diffusion process (right) to guide subsequent denoising steps toward semantically consistent video synthesis.}
    \label{fig:method}
    \vspace{-0.5cm}
\end{figure*}

In this section, we present our proposed framework for interpretable mid-generation correction in text-to-video diffusion models. 
We first formalize the diffusion-based generation process and introduce key notation (\S\ref{sec:background}). 
Then, we describe the two major components of our framework: the \textit{Semantic Assessment Supervisor} (\S\ref{sec:preview}) and the \textit{Semantic Modification Assistant} (\S\ref{sec:injection}), which jointly enable correction based on dynamic reasoning during inference.

\subsection{Background and Preliminaries}
\label{sec:background}

\subsubsection{Diffusion model}
Let a text-to-video diffusion model be parameterized by $\theta$, designed to generate a sequence of video frames conditioned on a textual prompt $p$. 
We denote the conditioning embedding as $c = E(p)$, where $E(\cdot)$ is a pretrained text encoder that maps the input prompt to a semantic feature space.

The model operates in the latent space, denoted by $\{x_t\}_{t=0}^T$, where $x_T \sim \mathcal{N}(0, I)$ represents Gaussian noise, and $x_0$ denotes the final clean latent to be decoded into a video sequence. 
Following the deterministic DDIM formulation~\cite{song2020denoising}, the sampling process is expressed as:
\begin{equation}
    x_{t-1} = \sqrt{\alpha_{t-1}}\, \hat{x}_0(x_t, t, c; \theta)
    + \sqrt{1 - \alpha_{t-1}}\, \epsilon_{\theta}(x_t, t, c),
    \label{eq:ddim}
\end{equation}
where $\epsilon_{\theta}$ is the denoising network that estimates the noise component at timestep $t$, and $\hat{x}_0$ is the model’s prediction of the corresponding clean latent.

After the final denoising step, the latent $x_0$ is decoded into a sequence of video frames by a video decoder $D(\cdot)$:
\begin{equation}
    \mathbf{v} = D(x_0) = \{v_1, v_2, \ldots, v_N\},
\end{equation}
where $N$ denotes the number of generated frames. 

\subsubsection{Incorporating multimodal reasoning}
To enable interpretable and adaptive correction during sampling, we integrate a MLLM denoted as $M(\cdot)$, which performs high-level semantic reasoning across both visual and textual modalities. 
At an arbitrary timestep $t$, we partially decode the latent $x_t$ through $D(\cdot)$ to obtain a coarse visual representation $\mathbf{v}_t = D(x_t)$, and feed it into the MLLM together with the textual prompt $p$:
\begin{equation}
    S_t = M(\mathbf{v}_t, p),
\end{equation}
where $S_t$ denotes the semantic reasoning feedback that encapsulates multimodal understanding signals and potential corrective signals.

\subsection{Semantic Assessment Supervisor}
\label{sec:preview}

A fundamental challenge of mid-generation correction lies in the nature of intermediate diffusion states.
At timestep $t$, the latent $x_t$ is dominated by stochastic noise and resides far from any interpretable visual manifold. 
Consequently, direct reasoning over $x_t$ offers little semantic insight, as its structure does not correspond to perceivable content.
This limitation requires a mechanism that can expose the evolving semantics of the model without disrupting its generative trajectory.

We draw inspiration from the diffusion process itself. 
Each denoising step implicitly carries an internal estimate of the underlying clean signal, formalized in DDIM as:
\begin{equation}
    \hat{x}_0(x_t, t, c; \theta)
    = \frac{x_t - \sqrt{1-\alpha_t}\,\epsilon_{\theta}(x_t, t, c)}{\sqrt{\alpha_t}}.
    \label{eq:predx0}
\end{equation}
Rather than treating $\hat{x}_0$ merely as an auxiliary computation, we reinterpret it as the model’s instantaneous hypothesis of the latent clean sample. 

Let $t_\text{start}$ and $t_\text{end}$ denote the evaluation interval boundaries and $\Delta$ the interval step.  
The set of timesteps at which the MLLM evaluation is performed is defined as:
\begin{equation}
\mathcal{T} = \bigl\{\, t \,\big|\, t = t_\text{start} + k \cdot \Delta,\ 
k = 0,1,\dots, \bigl\lfloor \tfrac{t_\text{end}-t_\text{start}}{\Delta} \bigr\rfloor \,\bigr\}.
\label{eq:interval_set}
\end{equation}

For each $t \in \mathcal{T}$, we decode an intermediate preview:
\begin{equation}
    \mathbf{v}_t^{pvw} = D(\hat{x}_0(x_t, t, c; \theta)),
    \label{eq:preview}
\end{equation}
which yields a low-fidelity, yet semantically aligned, representation that reflects the evolving generation state of the model. 

The intermediate visualization $\mathbf{v}^{pvw}_t$ and the textual prompt $p$ are then fed into the MLLM $M(\cdot)$ to produce diagnostic feedback:
\begin{align}
    S_t = (f_t, p_t^{+}, p_t^{-}) = M(\mathbf{v}_t^{pvw}, p), \quad
    \Delta c_t^{\pm} = E(p_t^{\pm}),
    \label{eq:correction}
\end{align}
where $f_t$ denotes the structured diagnostic feedback describing semantic inconsistencies (e.g., object incompleteness or attribute mismatch), 
$(p_t^{+}, p_t^{-})$ represents the positive and negative corrective prompts generated according to the diagnosis, 
and $E(\cdot)$ encodes these prompts into the conditioning embedding space.  
The resulting corrective embeddings $\Delta c_t^{\pm}$ are forwarded to the injection module (\S\ref{sec:injection}) for integration into subsequent diffusion updates.

In particular, before generating feedback $f_t$, our framework first evaluates the visual-semantic consistency between $\mathbf{v}_t^{pvw}$ and $p$. If no inconsistency is detected, the correction process is terminated early to avoid unnecessary semantic injection and reduce computational overhead.

\begin{algorithm}[H]
\caption{MLLM-Guided Semantic Correction for Text-to-Video Generation}
\label{alg:main}
\begin{algorithmic}
\STATE
\STATE {\textsc{INITIALIZE}}$(p, E, \epsilon_\theta, M, D)$
\STATE \hspace{0.5cm}$c \gets E(p)$
\STATE \hspace{0.5cm}$x_T \sim \mathcal{N}(0,I)$
\STATE \hspace{0.5cm}$\mathcal{T} \gets \{\, t_{\mathrm{start}} + k\Delta \mid k \in \mathbb{N},\ t \le t_{\mathrm{end}} \,\}$

\STATE
\STATE {\textsc{DIFFUSION SAMPLING}}
\STATE \hspace{0.5cm}\textbf{for} $t = T,\ldots,1$ \textbf{do}
\STATE \hspace{1.0cm}$\hat{x}_0 \gets (x_t - \sqrt{1-\alpha_t}\,\epsilon_\theta(x_t,t,c)) / \sqrt{\alpha_t}$
\STATE \hspace{1.0cm}$x_{t-1} \gets \sqrt{\alpha_{t-1}}\,\hat{x}_0 + \sqrt{1-\alpha_{t-1}}\,\epsilon_\theta(x_t,t,c)$

\STATE \hspace{1.0cm}\textbf{if} $t \in \mathcal{T}$ \textbf{then}
\STATE \hspace{1.5cm}{\textsc{Semantic Assessment}}
\STATE \hspace{1.5cm}$\mathbf{v}_t^{\mathrm{pvw}} \gets D(\hat{x}_0)$
\STATE \hspace{1.5cm}$(f_t,p_t^{+},p_t^{-}) \gets M(\mathbf{v}_t^{\mathrm{pvw}},p)$
\STATE \hspace{1.5cm}$\Delta c_t^{+} \gets E(p_t^{+}), \quad \Delta c_t^{-} \gets E(p_t^{-})$

\STATE \hspace{1.5cm}{\textsc{Semantic Correction}}
\STATE \hspace{1.5cm}$\lambda_t \gets \sqrt{1-\alpha_t} - \sqrt{\alpha_t/\alpha_{t-1}}\sqrt{1-\alpha_{t-1}}$
\STATE \hspace{1.5cm}$\tilde{x}_t \gets \sqrt{\alpha_t/\alpha_{t-1}}\,x_{t-1} + \lambda_t\,\epsilon_\theta(x_{t-1},t-1)$
\STATE \hspace{1.5cm}$\hat{x}_0 \gets (x_t - \sqrt{1-\alpha_t}\,\epsilon_\theta(\tilde{x}_t,t,\Delta c_t^{\pm})) / \sqrt{\alpha_t}$
\STATE \hspace{1.5cm}$x_{t-1} \gets \sqrt{\alpha_{t-1}}\,\hat{x}_0 + \sqrt{1-\alpha_{t-1}}\,\epsilon_\theta(\tilde{x}_t,t,\Delta c_t^{\pm})$
\STATE \hspace{1.0cm}\textbf{end if}
\STATE \hspace{0.5cm}\textbf{end for}

\STATE
\STATE {\textsc{OUTPUT}}
\STATE \hspace{0.5cm}$\mathbf{v} \gets D(x_0)$
\STATE \hspace{0.5cm}\textbf{return} $\mathbf{v}$
\end{algorithmic}
\end{algorithm}

\subsection{Semantic Modification Assistant}
\label{sec:injection}

Once the semantic feedback is obtained from the Semantic Assessment Supervisor, the key question becomes how to effectively inject this corrective information into the diffusion trajectory while maintaining temporal coherence and preserving model priors. 
To achieve this, we introduce a \textbf{Semantic Injection} process that performs controllable bidirectional refinement within the diffusion trajectory. 
This mechanism allows the model to partially reverse the accumulated semantic bias and then reintroduce corrected guidance under updated conditions. Putting the two components together, our full algorithm is presented in algorithm~\ref{alg:main}.

Generally, our semantic injection process consists of three sequential steps. 

\subsubsection{Semantic Dilution} Given the latent $x_{t-1}$ at step $t-1$, we first compute an intermediate latent $\tilde{x}_t$ by applying a one-step back transition that weakens the previously accumulated conditional influence:
\begin{equation}
\begin{aligned}
\tilde{x}_t &= 
\sqrt{\frac{\alpha_t}{\alpha_{t-1}}}\, x_{t-1} 
 + \lambda_t\,\epsilon_\theta(x_{t-1}, t-1, \phi), \\
 \lambda_t &= \sqrt{1-\alpha_t} - \sqrt{\frac{\alpha_t}{\alpha_{t-1}}}\sqrt{1-\alpha_{t-1}},
\label{eq:rollback}
\end{aligned}
\end{equation}
where $\epsilon_\theta$ denotes the noise prediction network parameterized by $\theta$, and $\phi$ represents the unconditional case. 
This step effectively dilutes the condition-induced semantic information, producing a semantically neutral latent $\tilde{x}_t$ that serves as a clean basis for subsequent feedback-guided correction.

\subsubsection{Semantic Injection} We apply the precomputed corrective embeddings $\Delta c_t^{\pm}$  as modified conditioning signals and perform a feedback-guided denoising step:
\begin{equation}
\begin{aligned}
\tilde{x}_{t-1} &= 
\sqrt{\alpha_{t-1}}\, \hat{x}_0(\tilde{x}_t, t, \Delta c_t^{\pm}; \theta) \\
&+ \sqrt{1 - \alpha_{t-1}}\, \epsilon_{\theta}(\tilde{x}_t, t, \Delta c_t^{\pm}),
\label{eq:forward1}
\end{aligned}
\end{equation}
where $\hat{x}_0(\tilde{x}_t, t, \Delta c_t^{\pm}; \theta)$ denotes the denoised estimate under the modified semantic guidance. 
This step injects the corrected semantic information into the diffusion trajectory, yielding a refined latent $\tilde{x}_{t-1}$ that incorporates high-level feedback from the preview representation.

\subsubsection{}{Trajectory Resumption} The corrected latent re-enters the normal diffusion trajectory and continues with normal denoising using the original conditioning:
\begin{equation}
\begin{aligned}
x_{t-2} &= 
\sqrt{\alpha_{t-2}}\, \hat{x}_0(\tilde{x}_{t-1}, t-1, c; \theta) \\
&+ \sqrt{1 - \alpha_{t-2}}\, \epsilon_{\theta}(\tilde{x}_{t-1}, t-1, c),
\label{eq:forward2}
\end{aligned}
\end{equation}
where $c$ is the initial condition. 
This step ensures that the generation process is resumed on the corrected trajectory.

In conclusion, this process enables the diffusion model to integrate high-level semantic feedback in a stable and interpretable manner. 

\section{Theoretical Foundations}
\label{sec:theory}

In this section, we provide a rigorous theoretical analysis of the proposed Semantic Assessment–Semantic Modification Assistant (SASMA). Our objective is to formally characterize how semantic feedback injection alters the diffusion trajectory and to establish conditions under which such modification yields a strictly smaller denoising error than standard DDIM inference.

We begin by explicitly formulating the three sequential operations that constitute one semantic correction cycle in SASMA. Based on these operations, we derive an exact four-term decomposition of the resulting update and subsequently analyze its denoising error relative to the baseline diffusion process.

\subsection{Semantic Correction as a Three-Step Diffusion Operator}

At a given timestep $t$, SASMA modifies the standard diffusion trajectory through the following three operations.

\paragraph{Semantic Dilution.}
The first step performs a controlled forward diffusion from $x_{t-1}$ to an intermediate noisy state $\tilde{x}_t$:
\begin{equation}
\begin{aligned}
\tilde{x}_t &= \sqrt{\frac{\alpha_t}{\alpha_{t-1}}}\, x_{t-1} \\
&+ \left(
   \sqrt{1-\alpha_t}
   - \sqrt{\frac{\alpha_t}{\alpha_{t-1}}}\sqrt{1-\alpha_{t-1}}
   \right)
   \epsilon_\theta(x_{t-1}, t-1, \phi).
\end{aligned}
\label{eq:step1}
\end{equation}
This step intentionally dilutes the current latent state, creating a re-noised representation that allows semantic feedback to exert non-local influence on the diffusion trajectory.

\paragraph{Semantic Injection.}
Given the intermediate state $\tilde{x}_t$, we perform a backward diffusion conditioned on the refined semantic guidance $\Delta c_t^{\pm}$:
\begin{equation}
\begin{aligned}
\tilde{x}_{t-1} &=
\sqrt{\alpha_{t-1}}\, \hat{x}_0(\tilde{x}_t, t, \Delta c_t^{\pm}; \theta) \\
&+ \sqrt{1 - \alpha_{t-1}}\, \epsilon_{\theta}(\tilde{x}_t, t, \Delta c_t^{\pm}).
\end{aligned}
\label{eq:step2}
\end{equation}
This step injects semantic corrections directly into the denoising process, yielding a refined latent $\tilde{x}_{t-1}$.

\paragraph{Trajectory Resumption.}
Finally, the diffusion trajectory is resumed under the original prompt condition $c$:
\begin{equation}
\begin{aligned}
x_{t-2} &=
\sqrt{\alpha_{t-2}}\, \hat{x}_0(\tilde{x}_{t-1}, t-1, c; \theta) \\
& + \sqrt{1 - \alpha_{t-2}}\, \epsilon_{\theta}(\tilde{x}_{t-1}, t-1, c).
\end{aligned}
\label{eq:step3}
\end{equation}

\subsection{Exact Algebraic Reformulation}

We first rewrite Eq.~\eqref{eq:step2} by substituting the definition of $\hat{x}_0$:
\begin{equation}
\begin{aligned}
\tilde{x}_{t-1}
&= \sqrt{\frac{\alpha_{t-1}}{\alpha_t}}\,\tilde{x}_t \\
&+ \left(
\sqrt{1-\alpha_{t-1}}
- \sqrt{\frac{\alpha_{t-1}(1-\alpha_t)}{\alpha_t}}
\right)\epsilon_{\theta}(\tilde{x}_t, t, \Delta c_t^{\pm}).
\end{aligned}
\label{eq:step2_simplified}
\end{equation}

Similarly, Eq.~\eqref{eq:step3} can be reformulated as:
\begin{equation}
\begin{aligned}
x_{t-2}
&= \sqrt{\frac{\alpha_{t-2}}{\alpha_{t-1}}}\,\tilde{x}_{t-1} \\
&+ \left(
\sqrt{1-\alpha_{t-2}}
- \sqrt{\frac{\alpha_{t-2}(1-\alpha_{t-1})}{\alpha_{t-1}}}
\right)\epsilon_{\theta}(\tilde{x}_{t-1}, t-1, c).
\end{aligned}
\label{eq:step3_simplified}
\end{equation}

Substituting Eq.~\eqref{eq:step2_simplified} into Eq.~\eqref{eq:step3_simplified} and expanding terms yields:
\begin{equation}
\begin{aligned} 
x_{t-2}
&= \sqrt{\frac{\alpha_{t-2}}{\alpha_t}}\,\tilde{x}_t \\ 
&+ \sqrt{\frac{\alpha_{t-2}}{\alpha_{t-1}}}\!
\left(\!
\sqrt{1-\alpha_{t-1}}\!
-\! \sqrt{\frac{\alpha_{t-1}(1-\alpha_t)}{\alpha_t}}\!
\right)\!
\epsilon_{\theta}(\tilde{x}_t, t, \Delta c_t^{\pm}) \\
&+ \left(
\sqrt{1-\alpha_{t-2}}
- \sqrt{\frac{\alpha_{t-2}(1-\alpha_{t-1})}{\alpha_{t-1}}}
\right)
\epsilon_{\theta}(\tilde{x}_{t-1}, t-1, c).
\end{aligned}
\label{eq:expanded}
\end{equation}

Substituting $\tilde{x}_t$ using Eq.~\eqref{eq:step1} and collecting terms, we arrive at the following four-term decomposition:
\begin{equation}
\begin{aligned}
x_{t-2}
&= \eta_1 x_{t-1}
+ \eta_2 \epsilon_\theta(x_{t-1}, t-1, \phi) \\
&+ \eta_3 \epsilon_{\theta}(\tilde{x}_t, t, \Delta c_t^{\pm})
+ \eta_4 \epsilon_{\theta}(\tilde{x}_{t-1}, t-1, c),
\end{aligned}
\label{eq:four_term}
\end{equation}
where the coefficients $\eta_1$--$\eta_4$ depend solely on the noise schedule:
\begin{align}
\eta_1 &= \sqrt{\frac{\alpha_{t-2}}{\alpha_{t-1}}}, \\
\eta_2 &= \sqrt{\frac{\alpha_{t-2}(1-\alpha_t)}{\alpha_t}}
- \sqrt{\frac{\alpha_{t-2}(1-\alpha_{t-1})}{\alpha_{t-1}}}, \\
\eta_3 &= \sqrt{\frac{\alpha_{t-2}(1-\alpha_{t-1})}{\alpha_{t-1}}}
- \sqrt{\frac{\alpha_{t-2}(1-\alpha_t)}{\alpha_t}}, \\
\eta_4 &= \sqrt{1-\alpha_{t-2}}
- \sqrt{\frac{\alpha_{t-2}(1-\alpha_{t-1})}{\alpha_{t-1}}}.
\end{align}

A key observation is that $\eta_2 + \eta_3 = 0$, which allows us to rewrite Eq.~\eqref{eq:four_term} as:
\begin{equation}
\begin{aligned}
x_{t-2}
&= \eta_1 x_{t-1}
+ \eta_4 \epsilon_{\theta}(\tilde{x}_{t-1}, t-1, c) \\
&+ \eta_3
\big[
\epsilon_{\theta}(\tilde{x}_t, t, \Delta c_t^{\pm})
- \epsilon_{\theta}(x_{t-1}, t-1, \phi)
\big].
\end{aligned}
\label{eq:compact}
\end{equation}

\subsection{Comparison with Standard DDIM}

For reference, the standard DDIM update from $t-1$ to $t-2$ is given by:
\begin{equation}
x_{t-2}
= \eta_1 x_{t-1}
+ \eta_4 \epsilon_{\theta}(x_{t-1}, t-1, c).
\label{eq:ddim_comparision}
\end{equation}

Comparing Eqs.~\eqref{eq:compact} and \eqref{eq:ddim}, we observe that SASMA introduces an additional semantic correction term that explicitly compensates for the discrepancy between the original and refined noise predictions.

\subsection{Denoising Error Analysis}

Let $\epsilon$ denote the true noise at timestep $t-1$. The denoising error of the standard DDIM update is:
\begin{equation}
\delta_{\text{DDIM}}
= |\eta_4| \cdot
\|\epsilon_{\theta}(x_{t-1}, t-1, c) - \epsilon\|.
\end{equation}

For SASMA, the denoising error becomes:
\begin{equation}
\begin{aligned}
\delta_{\text{SASMA}}
&= \big\|
\eta_4 [\epsilon_{\theta}(\tilde{x}_{t-1}, t-1, c) - \epsilon] \\
&+ \eta_3[
\epsilon_{\theta}(\tilde{x}_t, t, \Delta c_t^{\pm})
- \epsilon_{\theta}(x_{t-1}, t-1, \phi)]
\big\|.
\end{aligned}
\end{equation}

Define the state correction gain:
\begin{equation}
\Delta_{\text{state}}
= \|\epsilon_{\theta}(x_{t-1}, t-1, c) - \epsilon\|
- \|\epsilon_{\theta}(\tilde{x}_{t-1}, t-1, c) - \epsilon\|,
\end{equation}
and the semantic correction magnitude:
\begin{equation}
C_{\text{sem}}
= \|\epsilon_{\theta}(\tilde{x}_t, t, \Delta c_t^{\pm})
- \epsilon_{\theta}(x_{t-1}, t-1, \phi)\|.
\end{equation}

Applying the triangle inequality yields:
\begin{equation}
\delta_{\text{SASMA}}
\le |\eta_4|(\|\epsilon_{\theta}(x_{t-1}, t-1, c) - \epsilon\|
- \Delta_{\text{state}})
+ |\eta_3| C_{\text{sem}}.
\end{equation}

Therefore, when the enhanced semantic guidance satisfies:
\begin{equation}
|\eta_3| C_{\text{sem}}
< |\eta_4| \Delta_{\text{state}},
\end{equation}
we obtain:
\begin{equation}
\delta_{\text{SASMA}} < \delta_{\text{DDIM}}.
\end{equation}

This establishes that SASMA achieves a strictly smaller denoising error than standard DDIM under mild and practically satisfied assumptions, completing the theoretical justification.


\begin{table*}[t]
\centering
\caption{Quantitative comparison per dimension on VBench.}
\label{tab:vbench_main_per_dimension}
\renewcommand{\arraystretch}{1.1}
\setlength{\tabcolsep}{6pt}
\resizebox{\linewidth}{!}{
\begin{tabular}{l c c c c c c c c}
\toprule[1.5pt]
Model & Method & Subject Cons. $\uparrow$ & Aesthetic $\uparrow$ & Imaging $\uparrow$ & Human Act. $\uparrow$ & Spatial Rel. $\uparrow$ & Scene $\uparrow$ & Overall Cons. $\uparrow$ \\
\midrule
CogVideoX1.5 & Standard & 0.9088 & 0.5435 & 0.5624 & 0.8200 & 0.4328 & 0.3343 & 0.2483 \\
             & Ours (SASMA) & \textbf{0.9410} & \textbf{0.5633} & \textbf{0.5857} & \textbf{0.8440} & \textbf{0.4738} & \textbf{0.3894} & \textbf{0.2536} \\
\hline
HunyuanVideo & Standard & \textbf{0.9625} & 0.6012 & 0.6179 & 0.8840 & 0.5307 & 0.3581 & 0.2619 \\
             & Ours (SASMA) & 0.9604 & \textbf{0.6038} & \textbf{0.6323} & \textbf{0.9020} & \textbf{0.5803} & \textbf{0.3728} & \textbf{0.2643} \\
\hline
AnimateDiff & Standard & 0.9524 & 0.5938 & 0.6161 & 0.9240 & 0.5196 & 0.4987 & \textbf{0.2720} \\
            & Ours (SASMA) & \textbf{0.9569} & \textbf{0.6066} & \textbf{0.6463} & \textbf{0.9460} & \textbf{0.5355} & \textbf{0.5259} & 0.2713 \\
\toprule[1.5pt]
\end{tabular}
}
\end{table*}


\begin{table}[t]
\centering
\caption{Quantitative comparison of overall performance on VBench.}
\label{tab:vbench_main_total}
\renewcommand{\arraystretch}{1.1}
\setlength{\tabcolsep}{6pt}
\begin{tabular}{l c c c c}
\toprule[1.5pt]
Model & Method & Quality $\uparrow$ & Semantic $\uparrow$ & Total $\uparrow$ \\
\midrule
CogVideoX1.5 & Standard & 0.7717 & 0.6419 & 0.7457 \\
             & Ours (SASMA) & \textbf{0.7982} & \textbf{0.6689} & \textbf{0.7723} \\
\hline
HunyuanVideo & Standard & 0.8171 & 0.6975 & 0.7932 \\
             & Ours (SASMA) & \textbf{0.8201} & \textbf{0.7048} & \textbf{0.7970} \\
\hline
AnimateDiff & Standard & 0.8087 & 0.7195 & 0.7909 \\
            & Ours (SASMA) & \textbf{0.8136} & \textbf{0.7277} & \textbf{0.7963} \\
\toprule[1.5pt]
\end{tabular}
\end{table}


\section{Experiments}
\label{sec:experiments}

\subsection{Experimental Setup}

\subsubsection{Benchmarks and Evaluation Metrics}
We perform comprehensive experiments on two widely adopted text-to-video generation benchmarks: VBench~\cite{huang2023vbench} and ChronoMagic-Bench~\cite{yuan2024chronomagic}. 
VBench provides diverse and fine-grained evaluation dimensions, such as subject consistency, temporal coherence, and text-video alignment, while ChronoMagic-Bench focuses on evaluating generative temporal reasoning and long-horizon consistency.

\subsubsection{Comparison Baselines}
We compare our method with several state-of-the-art video diffusion frameworks, including CogVideoX1.5~\cite{yang2024cogvideox}, HunyuanVideo~\cite{kong2024hunyuanvideo} and AnimateDiff~\cite{guo2023animatediff}. 
All baselines are evaluated according to their default generation settings with prompts from the benchmark datasets. 
For each prompt, we generate multiple samples to measure the stability and consistency of the semantic correction performance.

\subsubsection{Implementation Details}
All experiments were performed on NVIDIA RTX 3090 GPUs. 
We adopt a DDIM-based sampling schedule with $T = 50$ steps. 
The semantic injection process begins with $t_s = 0.1T$ and ends at $t_e = 0.9T$, with an interval $\Delta = 5$ between the evaluation boundaries. 
For multimodal semantic reasoning, we employ VideoLLaMA3-7B~\cite{zhang2025videollama3frontiermultimodal} as MLLM. 
The model operates in a training-free inference setting, and all compared methods share identical random seeds and diffusion hyperparameters to ensure a fair comparison.

\subsubsection{Backbone-Specific Configurations}
All experiments adopt three representative text-to-video diffusion backbones. For each model, we denote the number of diffusion steps as $T$, the classifier-free guidance scale as $\gamma$, the number of generated frames as $N$, the frame rate as $\mathrm{FPS}$, and the resulting video duration as $D = N / \mathrm{FPS}$. For CogVideoX1.5, we set $(T,\gamma,N,\mathrm{FPS}) = (50, 6.0, 41, 8)$, yielding $D = 5\mathrm{s}$. For HunyuanVideo, we use $(T,\gamma,N,\mathrm{FPS}) = (30, 6.0, 41, 8)$, also producing $D = 5\mathrm{s}$. For AnimateDiff, due to its shorter temporal horizon, we adopt $(T,\gamma,N,\mathrm{FPS}) = (50, 7.5, 16, 8)$, resulting in $D = 2\mathrm{s}$. Our method operates in a training-free, plug-and-play manner and is applied consistently to all models without modifying any parameters or architectures. 

\begin{figure*}[!t]
    \centering
    \includegraphics[width=\linewidth]{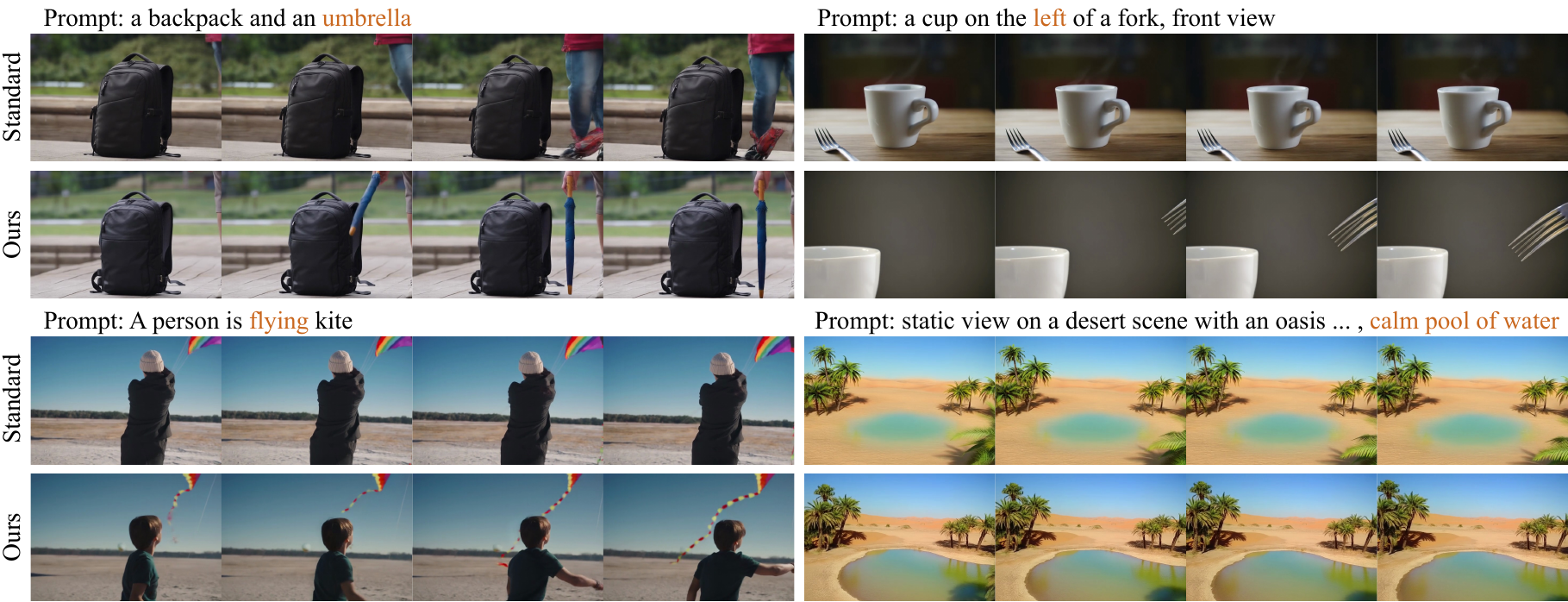}
    \caption{\textbf{Qualitative comparison on text-to-video generation.} Our method introduces an MLLM-guided feedback mechanism into the diffusion process, allowing mid-generation semantic correction without retraining. This integration enhances both temporal consistency and visual-semantic alignment in challenging cases such as multi-object composition, spatial relations, actions, and scenes. }
    \label{fig:case}
\end{figure*}

\subsection{Main Results}
\subsubsection{Quantitative Results}
\label{sec:quantitative}

Table~\ref{tab:vbench_main_per_dimension} and Table~\ref{tab:vbench_main_total} display the VBench results for three representative text-to-video backbones (CogVideoX1.5, HunyuanVideo and AnimateDiff). 
We present ten metrics in total: the first seven columns are per-dimension measures (Subject Consistency, Aesthetic Quality, Imaging Quality, Human Action, Spatial Relationship, Scene, and Overall Consistency), and the last three columns (Quality Score, Semantic Score, and Total Score) are aggregated scores computed from all per-dimension metrics.

The quantitative trends in Table~\ref{tab:vbench_main_total} demonstrate that our proposed SASMA improves overall performance across three architectures, with improvements spanning both semantic alignment and perceptual quality dimensions.
In particular, SASMA exhibits adaptive improvement behavior in different baseline strengths.
For weaker models such as CogVideoX1.5, the framework yields substantial gains in both structural and aesthetic aspects, while for already strong baselines like HunyuanVideo, it provides fine-grained refinements on semantic dimensions, especially in human action and spatial relationship understanding.
Moreover, the improvements are well balanced, as SASMA simultaneously increases both the Quality Score and Semantic Score across all architectures.
This indicates that a higher semantic correctness is achieved without compromising perceptual fidelity.
Although individual metrics for each dimension may exhibit minor variations, the aggregated Total Score consistently improves for the three models, confirming the robustness and generalizability of SASMA in diverse diffusion architectures.

\subsubsection{Qualitative Results}
\label{sec:qualitative}

\begin{figure}[t]
  \centering
    \includegraphics[width=1\linewidth]{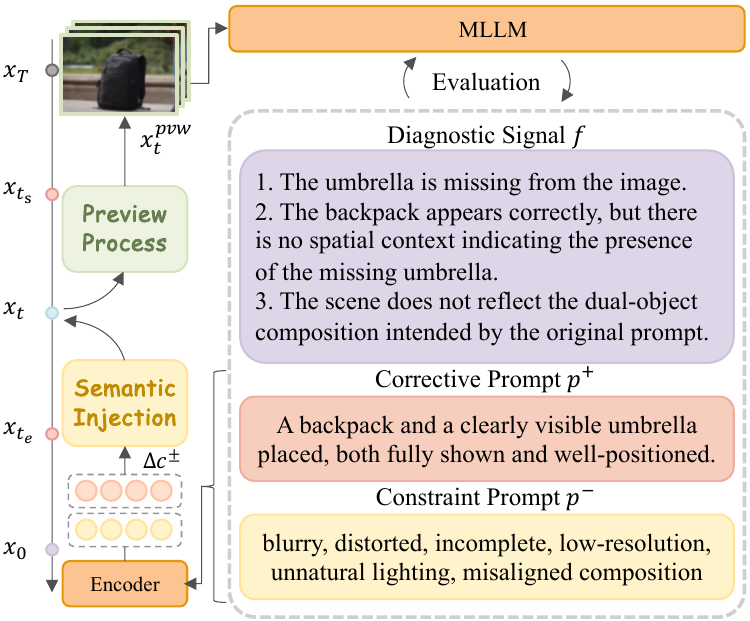}
    \caption{
        \textbf{Case of interpretable evaluation and correction.}
        Along the denoising trajectory from $x_T$ to $x_0$, our framework performs semantic assessment at discrete intervals on intermediate latents. 
        For each evaluated timestep, the MLLM produces three interpretable signals: a structured \textit{diagnostic signal} $f$ identifying semantic deviations, a \textit{corrective prompt} $p^{+}$ providing refined positive guidance, and a \textit{constraint prompt} $p^{-}$ specifying exclusion constraints. 
    }
    \label{fig:case_study}
    \vspace{-0.3cm}
\end{figure}

As shown in Figure~\ref{fig:case}, our proposed SASMA enhances both temporal consistency and visual-semantic alignment in text-to-video generation scenarios.
In addition, Figure~\ref{fig:case_study} presents a representative case of interpretable evaluation and correction, demonstrating how SASMA provides transparent and traceable feedback during the diffusion process.
The qualitative evaluation covers a variety of scenes, including multi-object arrangements, spatial relationships, dynamic human activities, and natural environments with multiple interacting elements, highlighting the robustness of SASMA across diverse content.

\subsection{Method Analysis and Ablation}
\label{sec:ablation}

\subsubsection{Ablations on the component of SASMA} We conduct ablation studies on Chronomagic-Bench-150 using CogVideoX1.5 to assess the contribution of each module in the proposed SASMA framework. 
As shown in Table~\ref{tab:ablation_cogvideox}, all modules consistently improve semantic and perceptual quality over the standard baseline.
The \textit{Semantic Injection} module enhances semantic coherence by embedding corrective signals into the diffusion trajectory. This process reduces semantic drift and improves frame-level consistency, demonstrating that even simple semantic steering can benefit mid-generation dynamics.
The \textit{Evaluation Module} further refines semantic accuracy through adaptive feedback based on MLLM assessment, improving overall alignment and structural correctness.
The \textit{Preview Module} contributes most to temporal stability. By generating cleaner intermediate previews, it enables the MLLM to reason over both spatial and temporal cues rather than noisy latents. This leads to more reliable feedback, especially for motion-related inconsistencies, resulting in smoother trajectories and stronger temporal coherence.


\begin{table}[t]
\centering
\caption{Ablation study of SASMA modules on Chronomagic-Bench-150 with CogVideoX1.5.}
\label{tab:ablation_cogvideox}
\renewcommand{\arraystretch}{1.05}
\setlength{\tabcolsep}{2.5pt}
\begin{tabular}{l c c c c}
\toprule[1.5pt]
Method & UMT-FVD $\downarrow$ & UMTScore $\uparrow$ & MTScore $\uparrow$ & CHScore $\uparrow$ \\
\midrule
Standard & 216.68 & 2.8485 & 0.3420 & 45.566 \\
+ Semantic Injection & 213.26 & 2.8619 & 0.3476 & 59.187 \\
+ Evaluation Module & \textbf{212.50} & \textbf{2.8771} & 0.3452 & 60.578 \\
+ Preview Module & 213.18 & 2.8653 & \textbf{0.3486} & \textbf{61.784} \\
\toprule[1.5pt]
\end{tabular}
\end{table}



\begin{figure*}[t]
\centering
\includegraphics[width=\linewidth]{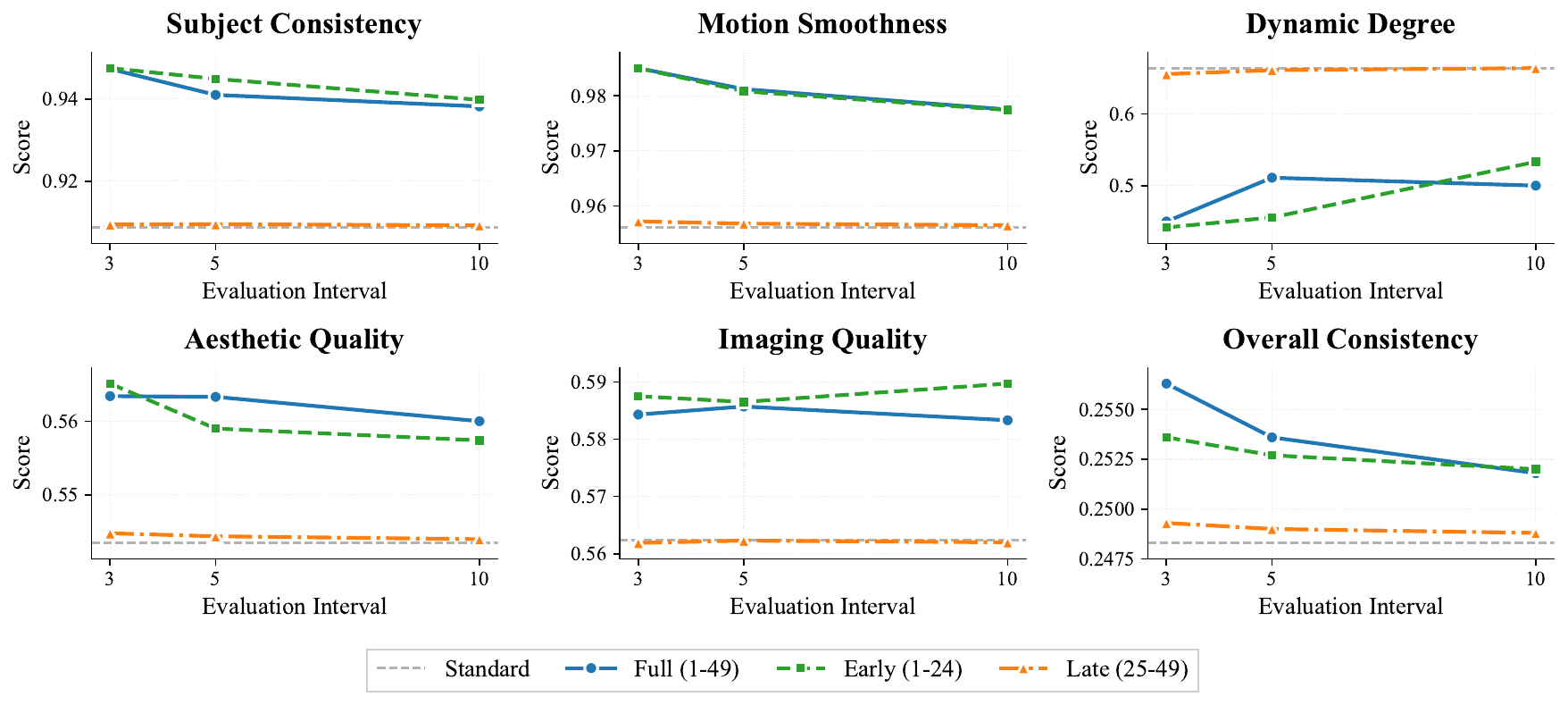}
\vspace{-0.7cm}
\caption{
\textbf{Effect of evaluation interval and stage scheduling.}
We compare three stage configurations (Early: 1-24, Late: 25-49, Full: 1-49) with three evaluation intervals (3, 5, 10) on CogVideoX1.5.
Early-stage high-frequency evaluation improves subject consistency and motion smoothness, while late-stage performance remains stable across frequencies.
Full-process mid-frequency scheduling (interval=5) achieves optimal balance.
The gray dashed line indicates baseline performance without SASMA.
}
\label{fig:hyperparameter}
\vspace{-0.4cm}
\end{figure*}

\begin{table*}[t]
\centering
\caption{Ablation on different numbers of evaluation--feedback rounds within the SASMA framework on CogVideoX1.5.}
\label{tab:ablation_rounds}
\renewcommand{\arraystretch}{1.05}
\setlength{\tabcolsep}{8pt}
\resizebox{\textwidth}{!}{
\begin{tabular}{l c c c c c c}
\toprule[1.5pt]
Method / Configuration & Motion Smooth. $\uparrow$ & Dynamic Deg. $\downarrow$ & Subject Cons. $\uparrow$ & Aesthetic $\uparrow$ & Imaging $\uparrow$ & Overall Cons. $\uparrow$ \\
\midrule
Standard & 0.9561 & \textbf{0.6639} & 0.9088 & 0.5435 & 0.5624 & 0.2483 \\
Ours (1 round) & 0.9812 & 0.4944 & 0.9396 & 0.5530 & 0.5745 & 0.2511 \\
Ours (2 rounds) & \textbf{0.9817} & 0.4167 & \textbf{0.9485} & \textbf{0.5668} & \textbf{0.6004} & 0.2531 \\
Ours (3 rounds) & 0.9812 & 0.5111 & 0.9410 & 0.5633 & 0.5857 & \textbf{0.2536} \\
\toprule[1.5pt]
\end{tabular}
}
\end{table*}


\subsubsection{Effect of Multi-round Evaluation Strategies}
We study how to schedule semantic feedback during inference, as the temporal ordering of corrective operations affects both efficacy and stability. 
We compare three configurations:
\begin{itemize}
  \item \textbf{One-round.} All operations (assessment, prompt polishing, and negative prompts) are applied in a single feedback round.
  \item \textbf{Two-rounds.} Round 1 performs semantic assessment, and Round 2 applies corrective actions (refined prompt and negative prompts).
  \item \textbf{Three-rounds.} Round 1 performs assessment, Round 2 applies prompt polishing, and Round 3 applies negative-prompt refinements.
\end{itemize}
Table~\ref{tab:ablation_rounds} demonstrates the results of CogVideoX1.5. 
The \textit{two-rounds} configuration achieves the best subject consistency and perceptual scores, while the \textit{three-rounds} approach yields higher overall consistency and more balanced improvements across temporal and semantic metrics. 
The \textit{one-round} scheme delivers substantial gains over the baseline, but is less effective in refining semantic fidelity.
These observations suggest that distributing corrective operations across multiple rounds enables progressive refinement, where each round builds upon the previous feedback, reducing abrupt over-corrections and cumulative errors. 

Specifically, the \textit{three-rounds} strategy separates prompt enhancement from negative prompt refinement, allowing the model to first correct positive semantic guidance before applying negative constraints. This staged approach prevents conflicts between complementary operations. 
Although the \textit{two-round} strategy achieves marginal gains in specific metrics, the \textit{three-round} strategy provides more consistent and balanced improvements with better generalization properties.
We adopt the \textit{three-round} configuration as our default based on its superior overall consistency and robustness. Notably, an early stopping mechanism is employed to terminate the refinement process once satisfactory semantic alignment is detected, allowing many samples to bypass additional correction rounds.

\subsubsection{Effect of Evaluation Interval and Stage Scheduling}
We investigate how the frequency of evaluation and the temporal scheduling affect the accuracy of the correction on the diffusion sampling trajectory.
Specifically, we systematically vary both the correction stage and the evaluation frequency throughout the 50-step sampling process. We compare three stage configurations (early: steps 1-24, late: steps 25-49, full: steps 1-49) with three evaluation intervals (3, 5, and 10), corresponding to high-frequency, mid-frequency, and low-frequency scheduling strategies, respectively.

Figure~\ref{fig:hyperparameter} presents the quantitative results in six VBench metrics on CogVideoX1.5.
The results reveal distinct stage-dependent patterns.
For the early stage (1-24), high-frequency evaluation (interval=3) consistently achieves the best performance in subject consistency, motion smoothness, and aesthetic quality, as this phase corresponds to the formation of semantic structures where frequent intervention helps stabilize scene layout and subject appearance.
In contrast, the late stage (25-49) exhibits minimal performance variation across different frequencies, indicating that the generation process has stabilized and excessive evaluation may introduce unnecessary perturbations.
For full-process evaluation (1-49), mid-frequency scheduling (interval=5) strikes an optimal balance between correction capability and computational efficiency, with performance approaching or even surpassing stage-specific strategies.

These observations suggest that, while a dense-to-sparse scheduling mechanism theoretically aligns with the progressive refinement nature of diffusion models, a fixed mid-frequency strategy across the entire sampling process achieves comparable effectiveness with simpler implementation.
The mid-frequency configuration maintains stable performance across all metrics, demonstrating robustness in balancing semantic correction and temporal consistency.
Based on these findings, we adopt interval=5 as our default configuration for subsequent experiments.

\subsubsection{Effect of MLLM Scale}

\begin{table*}[t]
\centering
\caption{Effect of MLLM scale on semantic correction quality.}
\label{tab:mllm_scale}
\renewcommand{\arraystretch}{1.05}
\setlength{\tabcolsep}{9pt}
\resizebox{\linewidth}{!}{
\begin{tabular}{l|cccccc|c}
\toprule[1pt]
MLLM & Motion Smooth. & Dynamic Deg. & Subject Cons. & Aesthetic & Imaging & Overall Cons. & VRAM (GB) \\
\midrule
Qwen2.5-VL-3B  & 0.9587 & \textbf{0.5018} & 0.9266 & 0.5345 & 0.5684 & 0.2467 & 9,290 \\
Qwen2.5-VL-7B  & \underline{0.9808} & \underline{0.4831} & \underline{0.9424} & \textbf{0.5669} & \textbf{0.5945} & \textbf{0.2528} & 17,142 \\
Qwen2.5-VL-72B & \textbf{0.9815} & 0.4228 & \textbf{0.9478} & \underline{0.5656} & \underline{0.5912} & \underline{0.2526} & 155,354 \\
\toprule[1pt]
\end{tabular}
}
\end{table*}

We investigate how the scale of the MLLM affects semantic correction quality by comparing Qwen2.5-VL at 3B, 7B, and 72B parameter scales.
Table~\ref{tab:mllm_scale} presents the results on CogVideoX1.5.

Larger models generally improve temporal and semantic consistency metrics. The 72B model achieves the highest motion smoothness and subject consistency, reflecting its superior capability in detecting fine-grained semantic misalignments. However, the 7B model outperforms 72B in perceptual quality metrics including aesthetic quality, imaging quality, and overall consistency.

Interestingly, the 3B model achieves the highest dynamic degree. We attribute this to the correction intensity: larger models tend to provide more detailed and stringent feedback, which can over-constrain the generation process and suppress motion dynamics. Smaller models apply more conservative corrections that preserve dynamic elements while still improving semantic alignment.

The 72B model requires approximately 9$\times$ the VRAM of the 7B model (155GB vs. 17GB), yet yields only marginal improvements in a subset of metrics. The 7B model strikes an effective balance between semantic understanding capability and correction intensity, achieving the best overall consistency while maintaining computational efficiency.

\subsection{Component-level Analysis}
\subsubsection{Effect of the Intermediate Preview Mechanism} 
\begin{figure*}[t]
    \centering
    \includegraphics[width=\linewidth]{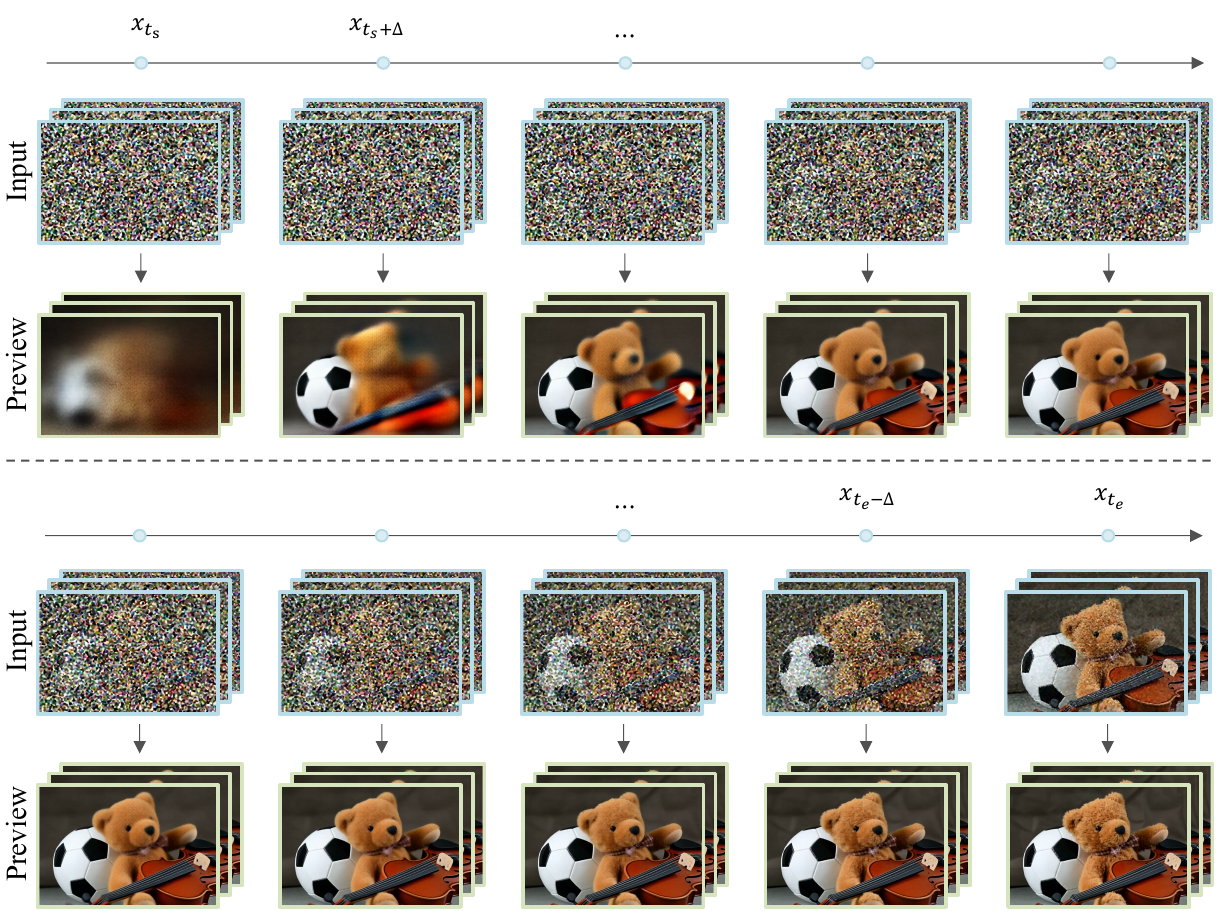}
    \caption{\textbf{Effect of Intermediate Previewing in Mid-Generation Evaluation.}
    From timestep $t_s$ to $t_e$, we insert preview operations at an interval of $\Delta$ steps along the diffusion trajectory.
    For each selected timestep $t$, we compare the noisy intermediate decoding $D(x_t)$ with the preview video obtained from $D({x}_t^{pvw})$. The prompt is "A teddy bear sitting between a football and a violin."}
    \label{fig:preview_case}
\end{figure*}

To illustrate the effect of our preview mechanism, Fig.~\ref{fig:preview_case} visualizes a sampling trajectory from timestep $t_s$ to $t_e$, where a preview is triggered at every $\Delta$ step. At each selected timestep $t$, we compare the raw decoded video $D(x_t)$ with the preview video obtained from $D({x}_t^{pvw})$. Without the preview mechanism, the MLLM is forced to evaluate sequences whose frames are still dominated by diffusion noise, making it difficult to reliably judge object presence, attributes, or actions. In contrast, the preview video provides a much cleaner and more semantically meaningful approximation of the final output, enabling the MLLM to more accurately identify semantic errors and produce actionable feedback for subsequent correction.

\subsubsection{Quantitative Validation of Preview Effectiveness}

To evaluate whether intermediate samples support reliable semantic assessment, we analyze MLLM judgments at different diffusion stages.
Experiments are conducted at 10 timesteps within a 50-step sampling process under two evaluation settings:
(1) \textit{Raw Decoding}, where noisy latents are directly decoded by the VAE, and
(2) \textit{Preview}, where the predicted clean latent $\hat{x}_0$ is decoded instead of the noisy $x_t$.

For each setting, we report two metrics:
(1) the \textit{agreement rate} between Qwen2.5-VL-7B and Qwen2.5-VL-72B, measuring judgment consistency, and
(2) the \textit{MATCH ratio}, defined as the proportion of samples judged as semantically aligned with the prompt.

\begin{figure*}[t]
    \centering
    \includegraphics[width=\textwidth]{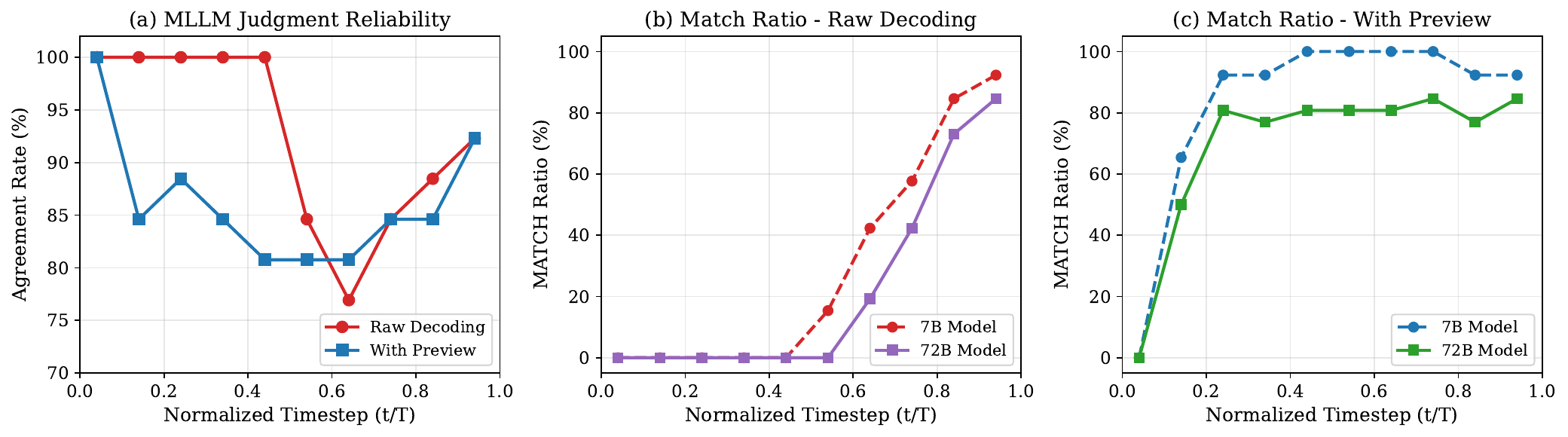}
    \caption{
        Quantitative analysis of mid-generation semantic evaluation.
        (a) Agreement rate between Qwen2.5-VL-7B and Qwen2.5-VL-72B.
        (b) MATCH ratio under raw decoding.
        (c) MATCH ratio with preview.
        Raw decoding produces near-zero MATCH predictions before $t/T \approx 0.5$, whereas preview enables earlier semantic recognition.
    }
    \label{fig:preview_analysis}
\end{figure*}

Figure~\ref{fig:preview_analysis}(a) shows that both settings reach high agreement ($>$85\%) at later timesteps.
However, high agreement alone does not indicate reliable semantic evaluation.
As shown in Figure~\ref{fig:preview_analysis}(b), raw decoding yields nearly zero MATCH predictions until $t/T \approx 0.5$, suggesting that intermediate samples remain semantically unrecognizable.

In contrast, preview-based evaluation produces valid semantic responses at much earlier stages.
As illustrated in Figure~\ref{fig:preview_analysis}(c), the MATCH ratio reaches 65\% at $t/T=0.14$ and exceeds 90\% after $t/T=0.24$.
These results indicate that preview significantly improves the visibility of semantic structure in intermediate samples, enabling earlier and more stable MLLM assessment.

\subsubsection{MLLM Feedback Quality Analysis}

\begin{figure*}[t]
    \centering
    \includegraphics[width=\linewidth]{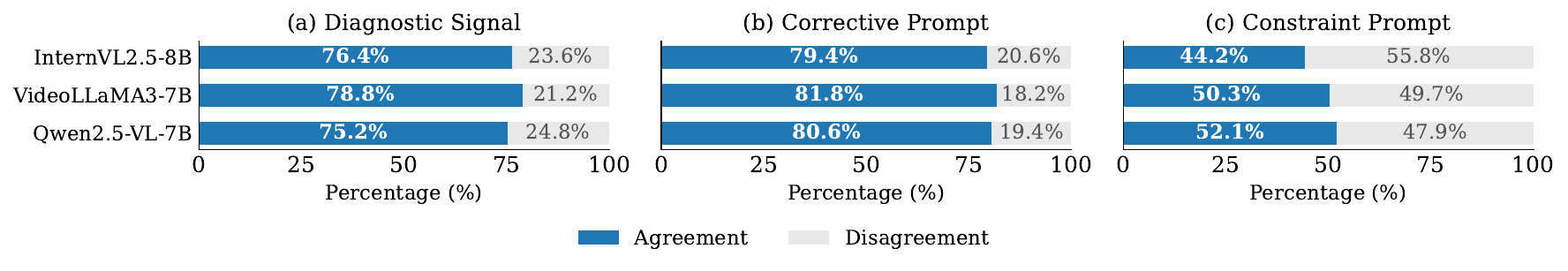}
    \caption{Evaluation of textual feedback quality across three models (Qwen2.5-VL-7B, VideoLLaMA3-7B, and InternVL2.5-8B). Each subplot shows the agreement rates between the model-generated feedback and the Qwen2.5-VL-72B verifier for (a) Diagnostic Signal, (b) Corrective Prompts, and (c) Constraint Prompt. The results are based on 165 test prompts from VBench. Darker bars indicate agreement (Yes), while lighter bars indicate disagreement (No).}
    \label{fig:feedback_quality}
\end{figure*}

To further assess the reliability of the textual signals produced by our method, we evaluate the quality of the generated \textit{Diagnostic Signal}, \textit{Corrective Prompt}, and \textit{Constraint Prompt} using an external verifier. Specifically, we generate these three forms of feedback using three different 7B or 8B MLLM models (Qwen2.5-VL-7B~\cite{qwen2.5-VL}, VideoLLaMA3-7B~\cite{zhang2025videollama3frontiermultimodal}, and InternVL2.5-8B~\cite{chen2024expanding}) and subsequently ask the more capable Qwen2.5-VL-72B~\cite{qwen2.5-VL} model to judge their consistency and correctness. The evaluation is conducted on 165 prompts selected from the \textit{subject consistency} and \textit{overall consistency} dimensions of VBench, which provide diverse scenarios to test the quality of textual feedback across different visual-text alignment challenges.

Figure~\ref{fig:feedback_quality} presents a comprehensive comparison of MLLM feedback quality in the three models. Across all models evaluated, the diagnostic signal and the corrective prompt consistently achieve high agreement rates with the Qwen2.5-VL-72B verifier, indicating that both the semantic descriptions of the mismatches and the proposed corrective guidance are generally reliable and well grounded in visual content. In contrast, the constraint prompt receives lower agreement scores, which suggests that describing what should be explicitly suppressed is inherently more ambiguous and model dependent. This observation aligns with the limitations discussed in Section~\ref{sec:limitation}, where we note that the effectiveness of the method may be partially contingent on the reliability of the MLLM feedback, especially in challenging cases or low-fidelity previews. Overall, the results confirm that the textual components driving our mid-generation correction framework are largely trustworthy, especially for diagnostic interpretation and positive guidance refinement.

\subsection{Efficiency and Inference Time Analysis}

\subsubsection{Inference Cost and Early Stopping Analysis}

\begin{table}[t]
\caption{Inference cost comparison under different configurations on CogVideoX1.5-5B with a single RTX 3090 GPU.}
\label{tab:inference_cost}
\centering
\renewcommand{\arraystretch}{1.05}
\setlength{\tabcolsep}{5pt}
\resizebox{\linewidth}{!}{
\begin{tabular}{lcccc}
\toprule[1pt]
Configuration & Overall Cons. & Time (s) & MLLM Calls & VRAM (GB) \\
\midrule
Standard        & 0.2483             & 254.43 & 0     & 13.7 \\
Ours (1-round)  & 0.2511             & 579.28 & 4.631 & 30.9 \\
Ours (2-rounds) & \underline{0.2531} & 588.95 & 3.289 & 30.9 \\
Ours (3-rounds) & \textbf{0.2536}    & 590.04 & 3.168 & 30.9 \\
\toprule[1pt]
\end{tabular}
}
\end{table}

\begin{table}[t]
\caption{Early stopping statistics on VBench using CogVideoX1.5-5B.}
\label{tab:early_stop}
\centering
\renewcommand{\arraystretch}{1.05}
\setlength{\tabcolsep}{11pt}
\resizebox{\linewidth}{!}{
\begin{tabular}{lcccc}
\toprule[1pt]
Dataset & Stop $\leq$1 & Stop $\leq$2 & Stop $\leq$3 & Avg. Calls \\
\midrule
VBench & 34.96\% & 66.95\% & 78.79\% & 3.168 \\
\toprule[1pt]
\end{tabular}
}
\end{table}

Table~\ref{tab:inference_cost} reports the computational cost under different multi-round configurations. Compared with the standard diffusion process, introducing semantic feedback increases the runtime due to the additional preview generation, MLLM inference, and semantic injection operations. The peak memory usage rises to 30.9,GB because the MLLM is loaded during inference, while the memory footprint remains constant across different configurations.
Although the two-round and three-round strategies decompose the correction process into multiple stages, their runtime is only slightly higher than that of the one-round setting. Meanwhile, the average number of MLLM calls decreases as more stages are introduced.

This behavior is explained by the early stopping mechanism. During sampling, semantic evaluation is performed at scheduled timesteps (every $\Delta$ steps). Before executing the full self-reflection pipeline, the MLLM is first asked whether the current video already matches the original prompt. If semantic alignment is confirmed, all subsequent correction operations at later timesteps are skipped and the generation continues with the standard diffusion process. Otherwise, the corresponding staged operations (assessment, prompt refinement, or negative prompt refinement) are executed according to the selected configuration.

Table~\ref{tab:early_stop} further reports the early stopping distribution on VBench. Notably, 34.96\% of samples terminate after a single MLLM evaluation, and 66.95\% stop within two calls. Overall, 78.79\% of samples converge within three calls, well below the maximum number of evaluations allowed by the sampling schedule. The average number of calls is only 3.168, indicating that most samples achieve semantic alignment at early stages.

As a result, the one-round configuration applies all corrections simultaneously at each evaluation step, which can lead to over-correction and introduce new semantic inconsistencies that require additional evaluations to resolve. In contrast, multi-round configurations apply corrections progressively, allowing finer-grained adjustments that are less likely to overshoot. This leads to faster convergence and fewer MLLM calls on average. The mechanism explains why the runtime difference between configurations is small despite their increased structural complexity. In large-scale generation scenarios, early stopping further reduces the amortized inference cost.

\subsubsection{Component-wise Efficiency Breakdown}

\begin{table}[t]
\centering
\caption{Component-wise efficiency breakdown of SASMA on CogVideoX1.5-5B.}
\label{tab:efficiency_breakdown}
\renewcommand{\arraystretch}{1.1}
\setlength{\tabcolsep}{6pt}
\resizebox{\linewidth}{!}{
\begin{tabular}{lccc}
\toprule[1pt]
Component & Avg. Time (s) & Total Time (s) & Percentage \\
\midrule
MLLM Inference & 7.53 & 75.28 & 19.68\% \\
Intermediate Preview & 14.76 & 147.59 & 38.59\% \\
Semantic Injection & 15.96 & 159.58 & 41.73\% \\
\hline
Total & — & 382.45 & 100\% \\
\toprule[1pt]
\end{tabular}
}
\end{table}

Table~\ref{tab:efficiency_breakdown} reports the time distribution of the major components in SASMA, measured over 10 semantic evaluation steps on CogVideoX1.5-5B. 
The results show that the additional overhead is dominated by the latent-space operations associated with semantic injection (41.73\%) and intermediate preview generation (38.59\%), while MLLM inference accounts for a relatively smaller portion (19.68\%).
This observation indicates that the computational cost of SASMA is primarily determined by video decoding and diffusion-related processing rather than the multimodal reasoning itself. In practice, this suggests that further acceleration should focus on improving preview efficiency (e.g., low-resolution decoding or lightweight decoders) and reducing the frequency of semantic injection. 
Although SASMA introduces additional computation, the overhead remains moderate compared with the overall diffusion cost. Combined with the performance gains reported in Table~\ref{tab:vbench_main_per_dimension}, this trade-off demonstrates the practical feasibility of mid-generation semantic correction.

\section{Conclusion}
\label{sec:conclusion}


We present a training-free framework for mid-generation semantic correction in text-to-video diffusion models. By decoupling semantic interpretation from trajectory manipulation, the Semantic Assessment Supervisor generates coherent intermediate previews and structured diagnostic signals, while the Semantic Modification Assistant performs trajectory corrections without modifying model parameters. This design enables precise semantic control and consistently improves semantic accuracy on diverse prompts. Our work demonstrates that semantic coherence can be enhanced using external multimodal reasoning signals during sampling, without modifying architectures or task-specific training, and we believe this perspective generalizes to other conditional synthesis tasks.

\section{Limitations}
\label{sec:limitation}
Our framework relies on the semantic assessment capability of the MLLM, which may exhibit prejudice when evaluating certain attributes or object categories in the low-fidelity intermediate previews. Although such limitations do not fundamentally undermine the effectiveness of the approach, they could lead to suboptimal corrections in edge cases where the diagnostic signals of MLLM are untrustworthy, suggesting that the robustness of the method is partially contingent upon the quality and generalization capability of the underlying MLLM.




\bibliographystyle{IEEEtran}
\bibliography{reference}

\end{document}